%% file: main.tex
\documentclass[runningheads]{llncs}

\usepackage{eccv}

\usepackage{eccvabbrv}

\usepackage{graphicx}
\usepackage{booktabs}

\usepackage[utf8]{inputenc}
\usepackage{tcolorbox}
\usepackage{xcolor}
\usepackage{geometry}
\usepackage{lmodern}

\definecolor{boxbg}{RGB}{245, 248, 255}
\definecolor{boxtitle}{RGB}{30, 80, 160}
\definecolor{boxframe}{RGB}{70, 130, 210}
\tcbuselibrary{skins, breakable}

\usepackage[accsupp]{axessibility}  

\usepackage{hyperref}

\usepackage{orcidlink}

\usepackage{booktabs}
\usepackage{colortbl}
\usepackage{xcolor}
\usepackage{multirow}
\usepackage{adjustbox}
\usepackage{enumitem}
\usepackage{wrapfig}

\usepackage{pifont}
\usepackage{amssymb}
\usepackage{caption}

\usepackage{empheq}

\newcommand{\lxcapvs}{\vspace*{-0.8em}}
\newcommand{\lxtailvs}{\vspace*{-0.1em}}

\begin{document}


\title{RegRet: Enhancing Region-Level Retrieval in Large Multimodal Models} 

\titlerunning{RegRet: LMMs for Region-level Retrieval}

\author{Xun Liang\inst{1} \and
Honghui Yang\inst{2} \and
Weihang Pan\inst{3} \and
Ruisi Zhao\inst{1} \and
Boyuan Pan\inst{2}\thanks{Corresponding author: \email{panby@zju.edu.cn}} \and
Yao Hu\inst{2} \and
Wenxiao Wang\inst{3} \and
Binbin Lin\inst{3}\thanks{Corresponding author: \email{binbinlin@zju.edu.cn}} \and
Deng Cai\inst{1}}

\authorrunning{X.~Liang et al.}

\institute{State Key Lab of CAD\&CG, Zhejiang University \and
Xiaohongshu Inc. \and
School of Software Technology, Zhejiang University}

\maketitle

\begin{abstract}
Region-level retrieval aims to align user-specified image regions with relevant regions or textual descriptions, playing a crucial role in realworld applications such as e-commerce product search and RAG. Although recent Large Multimodal Models (LMMs) have made significant strides in multimodal retrieval, they primarily focus on global-level tasks and struggle to capture effective region-level representations.
To bridge this gap, we present \textbf{\textit{RegRet}}, an LMM-based \textbf{Reg}ion-level \textbf{Ret}rieval framework that enhances the regional representations without compromising overall global retrieval performance. At its core, RegRet integrates a Region‑Aware Encoder to capture detailed regional features while balancing them with the global background context. 
To further enhance the fine-grained understanding and discriminability of representations, we design a multi-stage training pipeline that includes detailed localized captioning and regional contrastive learning tasks. 
In addition, considering the absence of region-level contrastive training data and the limited diversity of evaluation tasks in current benchmarks, we introduce the \textbf{\textit{REGMB}} benchmark. It comprises 225k contrastive pairs, covering four multimodal retrieval tasks.
Extensive experiments validate the effectiveness of our approach. RegRet outperforms strong baselines in the zero-shot setting. Further training with contrastive learning leads to an average improvement of more than 20\% on both REGMB and public benchmarks, while achieving comparable or better results on global-level retrieval tasks. 
\end{abstract}

\section{Introduction}
\label{sec:intro}

\begin{wrapfigure}{r}{0.5\textwidth}
    \centering
    \vspace{-2em} 
    \includegraphics[width=0.48\textwidth]{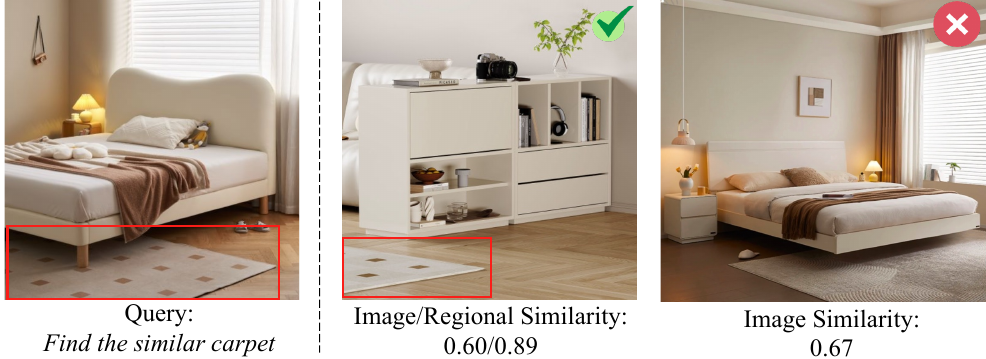}
    \vspace{-0.5em} 
    \caption{The difference between region-level and global-level retrieval. Without the ROI, the negative candidate will be retrieved.}
    \label{fig:bedroom}
    \vspace{-4em} 
\end{wrapfigure}

In recent years, multimodal information retrieval has been applied across a variety of domains. In many scenarios, such as e-commerce product search~\cite{jin2023eclip,dong2022m5product,zhang2018visual}, medical image retrieval~\cite{lee2023medical}, and retrieval-augmented generation~\cite{kim2026pixelgroundedretrievalknowledgeablelarge}, modern industrial retrieval systems typically leverage region-of-interest (ROI) in queries and candidates to enhance region-level matching. 
For example, as illustrated in~\cref{fig:bedroom}, a shopper may upload a bedroom photo together with a bounding box to find carpets with similar fine-grained texture. Without the ROI, a room that shares the same layout will be assigned a higher similarity score. Although existing studies primarily focus on building image-level representations and assume that ROIs are hard to acquire, a practical question remains underexplored: \textbf{How to push the boundaries of retrieval performance when region-level information is available?}

Despite the industrial practical value of the above question, existing multimodal retrieval models~\cite{liu2025lamra,chen2025mmE5,lin2024mmembed} are not designed for region-level tasks and thus face two challenges: 
\textbf{(1) Balancing the regional information and background context.} Although some visual understanding LMMs~\cite{zhang2025gpt4roi,guo2024regiongpt,wang2025grasp} leverage naive prompting strategies (\eg, crop, ROIAlign) to handle region inputs, directly using them for retrieval will lead to either excessive focus on local regions or the background becoming a distractor for retrieval, leading to suboptimal results as shown in~\cref{sec:case-study,sec:abla-prompt}. \textbf{(2) Comprehensive training and evaluation data.} There are no large-scale region-level contrastive datasets to support training. Meanwhile, existing evaluation benchmarks are also scarce. Most of them are centered on image-to-image task, with limited scenario coverage.

\begin{figure*}[!t]
    \centering
    \includegraphics[width=\linewidth, keepaspectratio]{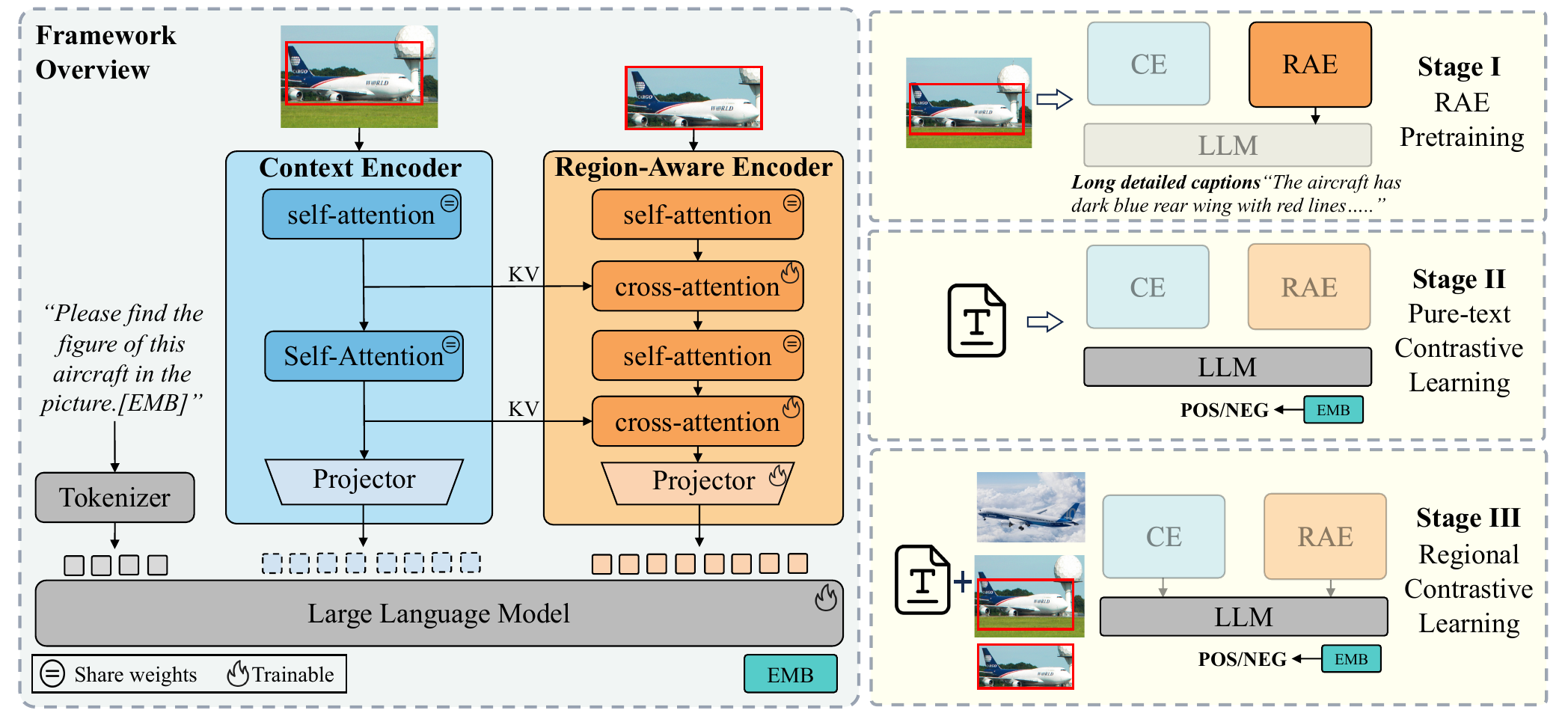}
    \caption{
    Framework overview of RegRet (\S~\ref{sec:methods}). The image and its ROI are processed by the native visual encoder (CE) and the RAE, respectively. Each RAE layer applies cross-attention with CE's corresponding layer to extract region-specific signals at matching semantic levels. CE tokens in the dashed box can be used in isolation for global-level retrieval, or jointly used with RAE tokens. 
    An LLM integrates the multimodal tokens and produces the final embedding for retrieval.
    }
    \label{fig:mainfig}
    \lxtailvs
    \vspace{-1em} 
    
\end{figure*}

To address the first challenge, we present \textbf{\textit{RegRet}}, an LMM-based framework for \textbf{Reg}ion-level multimodal \textbf{Ret}rieval that also supports global-level tasks. At its core, RegRet integrates a Region-Aware Encoder (RAE) alongside the native vision encoder (denoted as context encoder, CE) to build regional embeddings and balance background information. As illustrated in~\cref{fig:mainfig}, RAE utilizes cross-attention with global-level features from CE to selectively include necessary background context. The attention modules are organized into a layer-wise paradigm with CE, allowing it to gather information at different semantic levels from all ViT layers. 
The visual tokens from RAE and CE are aligned with LLM separately, so that it can preserve the image-level retrieval performance after training on region-level datasets. 
Furthermore, to boost RegRet’s retrieval performance and take full advantage of the properties of RAE, we devise a three-stage training pipeline, including (1) RAE pretraining, which employs the Detailed Localized Captioning task~\cite{lian2025dam} under the next-token prediction paradigm to encourage the model to learn detailed regional representations and necessary background context; (2) Pure-text contrastive learning, which use text-only pairs to convert the LMM's language generation capability into embedding capability; and (3) Regional contrastive learning, which use region-level contrastive pairs to further improve the retrieval performance.

To address the second challenge, we establish \textbf{\textit{REGMB}}, a \textbf{Reg}ional \textbf{M}ulti-modal retrieval \textbf{B}enchmark. It covers four multimodal retrieval tasks~\cite{zhang2024gme} and supports ROI-based training and evaluation.  REGMB contains 225k region-level contrastive pairs, primarily derived from a manually curated and privacy-sanitized subset from the social media community. The contrastive pairs are first retrieved by SigLip2~\cite{tschannen2025siglip} and filtered by human annotators. It is further enriched with automatically annotated open-source datasets~\cite{kakaobrain2022coyo-700m,kirillov2023segany,vismin2024,jiao2024imgdiffcontrastivedatasynthesis}.
Combining the above designs, RegRet outperforms all strong baselines even without additional region-level contrastive training. Specifically, it exceeds the average performance of LMMs by 14.3\% on REGMB and 15.0\% on existing public benchmarks. When fine-tuned with REGMB, its region-level retrieval performance is further boosted, achieving 7.4\% and 7.8\% gains over its zero-shot counterpart on the two benchmarks, respectively. Additionally, RegRet delivers on-par or even superior performance to state-of-the-art LMMs on M-BEIR, the global-level retrieval benchmark. Our contributions are as follows:

\begin{itemize}[leftmargin=*, itemsep=0em] 
\item We propose RegRet, an LMM that incorporates a Region-Aware Encoder and a three-stage training pipeline, thereby significantly enhancing region-level retrieval capabilities.
\item We construct REGMB, the first comprehensive region-level multimodal retrieval benchmark covering four typical scenarios, providing the necessary training and evaluation data that are absent in existing datasets.

\item We conduct extensive experiments on REGMB and public benchmarks. RegRet achieves leading region-level performance in both zero-shot and fine-tuned settings, while also preserving global-level retrieval capability.
\end{itemize}

\section{Related Works}
\label{Sec:related-work}
\textbf{LMMs for Multimodal Embedding.} 
Recent advances in LLMs have demonstrated remarkable performance in embedding learning. Early efforts, such as E5~\cite{wang2022e5} and NV-Embed~\cite{lee2024nvembed}, adapted the generative capabilities of large language models to text retrieval tasks. Building upon this, LamRA~\cite{liu2025lamra}, Vlm2Vec~\cite{jiang2024vlm2vec}, E5-V~\cite{jiang2024e5v}, and MMEMBED~\cite{lin2024mmembed} extended the paradigm from text retrieval to multimodal retrieval.
More recently, MME5~\cite{chen2025mmE5} leveraged synthetic data, while RzenEmbed~\cite{jian2025rzenembed} adopted a refined InfoNCE loss to enhance retrieval performance. 
Compared to traditional approaches, LMM-based methods exhibit stronger representation and generalization abilities. 

\par \noindent \textbf{Region-Level Representation Learning.}
A growing body of work focuses on instance-level alignment between image regions and text. RegionCLIP~\cite{zhong2022regionclip}, FGCLIP~\cite{xie2025fgclip}, and FineCLIP~\cite{jing2024fineclip} crop regional features via ROIAlign and align them to fine-grained text descriptions, while LongCLIP~\cite{zhang2024long} and DreamCLIP~\cite{zheng2024dreamlip} adopt long captions to achieve finer-grained semantic grounding between image regions and text.
However, LMM-based region-level retrieval methods remain underexplored. 
Some LMMs designed for visual understanding propose some regional prompting strategies, yet these cannot be directly adapted to retrieval tasks. For example, GPT4ROI~\cite{zhang2025gpt4roi}, RegionGPT~\cite{guo2024regiongpt}, and GRASP~\cite{wang2025grasp} use ROIAlign to crop regional features and concatenate them in the context as auxiliary images. 
However, naive background cropping may lead to misinterpretation of the region when the ROI is small, while using auxiliary images tends to introduce irrelevant parts of the original image as distractor noise in the final embedding.
DAM~\cite{lian2025dam} uses an encoder-decoder-like localized ViT for target-region encoding, but directly adapting it to retrieval tasks leads to conflicts between regional and global features, preventing the model from leveraging data of both types to fully unlock retrieval performance.

\par \noindent \textbf{Multimodal Information Retrieval Benchmarks.}
Representative global-level benchmarks include MMEB~\cite{jiang2024vlm2vec}, M-BEIR~\cite{wei2023uniir}, and MMEB-V2~\cite{meng2025vlm2vec2}, which cover a wide range of fused-modal tasks (\eg, image-text-to-image retrieval) and extensive training data. In contrast, region-level benchmarks remain limited. A widely adopted evaluation is to reformulate the box classification task of COCO~\cite{lin2014microsoft} as an image-to-text retrieval problem~\cite{xiao2025flair, zhong2022regionclip}, as shown in~\cref{fig:REGMB-case}. Yet this simplified text-matching paradigm falls short of modern retrieval system requirements. Some early attempts focus on image-to-image retrieval in given scenarios, such as $\mathcal{R}$Oxford~\cite{radenovic2018revisitingoxford} for landmark matching and DeepFashion2~\cite{DeepFashion2} for consumer-to-in-shop clothes retrieval. More recently, ILIAS~\cite{ilias} introduces an instance-level benchmark supporting both regional image-image and image-text retrieval. Nevertheless, existing benchmarks are confined to specific cross-modal tasks and do not support training. To date, a comprehensive region-level benchmark that supports training and covers diverse tasks remains absent.

\section{Methods}
\label{sec:methods}

In this section, we first formulate the region-level multimodal information retrieval task in~\cref{sec:methods-task-fomu}. Then we elaborate on RegRet in~\cref{sec:methods-model-design}, including the way to handle region-level prompts effectively and the architecture of the Region-Aware Encoder. Finally, in~\cref{sec:methods-train-pipe}, we describe the three-stage training strategy designed to boost RegRet's retrieval performance.

\subsection{Task Formulation}
\label{sec:methods-task-fomu}
Region-level multimodal retrieval aims to find the most relevant image or text at the region level. Formally, we define the query set as $Q = \{(q_1, r_{q_1}), (q_2, r_{q_2}), \ldots,$ $ (q_M, r_{q_M})\}$, 
where each query $q_k$ may consist of an image, a text description, or a combination of both. If the query includes an image, it is associated with an ROI $r_{q_k}$; otherwise, the $r_{q_k}$ is empty. We denote by the operator $q\otimes r_q$ the process of constructing a region-specific query, where the image part of $q$ is cropped to the ROI $r_q$ and the text part remains unchanged. Similarly, we define the candidate set as $C = \{(c_1, r_{c_1}), (c_2, r_{c_2}), \ldots, (c_N, r_{c_N})\}$, where each $c_k$ comprises images, text, or interleaved formats. 
Given a query $q_k$, the retrieval process selects the candidate $c_* \in C$ whose regional representation $c_* \otimes r_{c_*}$ is most semantically aligned with $q_k \otimes r_{q_k}$, denoted as:
\begin{equation}
c_* = argmax_{\{c\}\in C}[\langle\Phi(q_k\otimes r_{q_k}),\ \Phi(c\otimes r_{c})\rangle],
\end{equation}
where $\Phi(\cdot)$ denotes the function that embeds the query and candidates into vector representations, and $\langle\cdot,\cdot\rangle$ denotes the similarity function, such as the cosine similarity. In practice, target ROIs can be image patches ~\cite{yao2021filip} or region proposals from object detectors~\cite{zhong2022regionclip,dong2022m5product,li2021unimo}.

\subsection{Model Design}
\label{sec:methods-model-design}
The overall framework of RegRet is shown in~\cref{fig:mainfig}. It is built based on an LMM. First, the context encoder processes the entire image query into visual tokens, capturing holistic image content. The region-aware encoder then refines the middle-layer features of the context encoder to extract visual tokens that contain fine-grained information and necessary background context about the ROI. Finally, the LLM backbone integrates these multimodal input tokens into embeddings for retrieval.

\subsubsection{Regional Visual Prompts}
Since LMMs do not support regional inputs, prompting regions to it may largely impact the performance. Though there are several strategies, each of them has inherent flaws for retrieval tasks: (1) \textbf{Visual mark.} As shown in~\cref{fig:framework-abla}(a), it renders marks~\cite{yang2023setofmark} over the image to indicate the region. This method relies on visual grounding capability, thus sometimes it cannot sufficiently highlight the region, and background information still dominates. 
(2) \textbf{Crop.} Cropping the image region (or visual feature through ROIAlign) may discard the necessary background that helps the model to interpret the region correctly. For example, in Task 1 of~\cref{fig:REGMB-case}(b), the model will mistake the umbrella for white cloth without the background of the shops. 
(3) \textbf{Auxiliary image.} Regions are considered as independent images and concatenated to the tail of the original image as input~\cite{yu2025vpt,wang2025grasp}, as depicted in~\cref{fig:framework-abla}(b). However, the model tends to pay excessive attention to the original image, thus introducing distractor components into the final embedding. A more detailed case study of the above strategies can be found in~\cref{sec:case-study}.

To overcome the shortcomings of the above approaches, we adopt a different strategy: regional inputs are cropped and encoded via a separate Region-Aware Encoder (RAE), as shown in~\cref{fig:framework-abla}(d). By designing the architecture and training strategy of RAE, we enable it to adaptively balance background context and regional information using learnable parameters, thereby addressing the limitations of naive prompting strategies. The visual tokens from CE are marked in a dashed box, indicating that they are independent of the RAE and can be discarded, leveraged as auxiliary tokens for RAE in regional retrieval, or used in isolation for global-level retrieval.

\subsubsection{Region-Aware Encoder}

\begin{figure*}[!t]
    \centering
    \includegraphics[width=\linewidth, keepaspectratio]{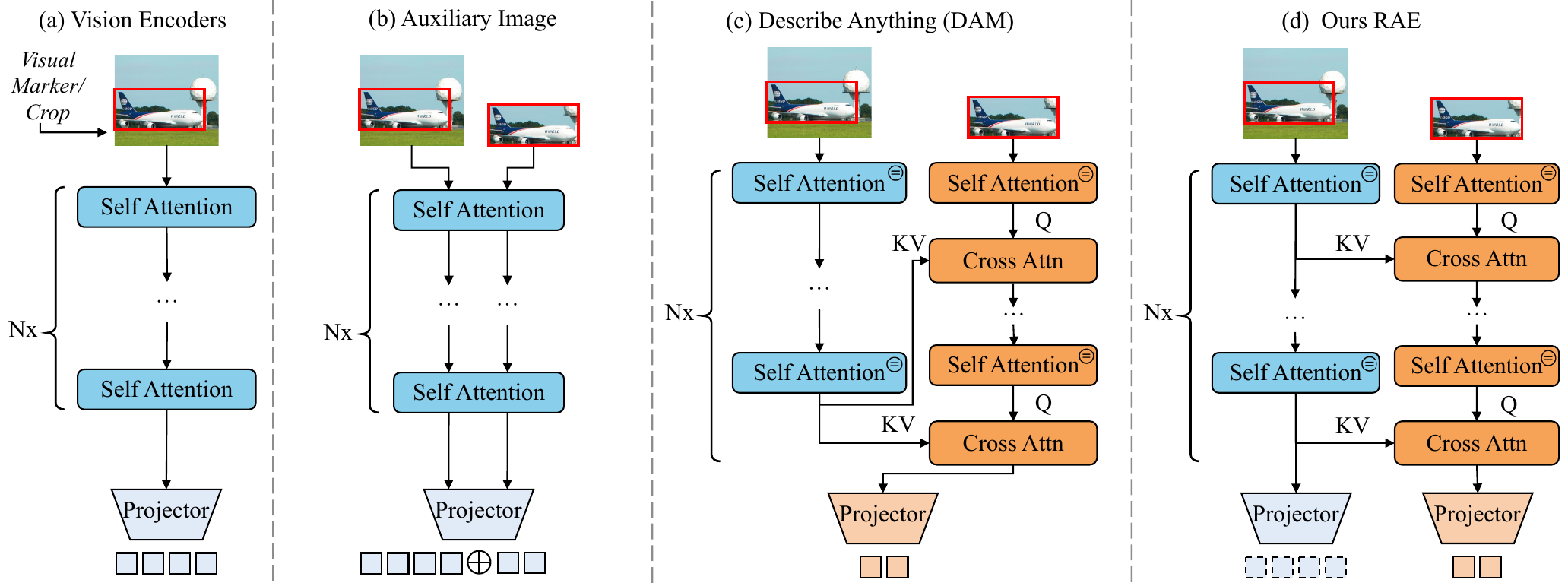}
    \caption{Comparison of vision backbone architectures and regional prompting strategies in previous methods (\S~\ref{sec:methods-model-design}). (a) Vanilla ViT with the visual mark or cropping. (b) Auxiliary Image concatenates the target region as an independent image. (c) DAM extracts regional features from the global-level visual embeddings of the last layer. Both ViTs share the same projector. (d) Our RAE adopts separate projectors and layer-wise coordination to balance the regional and global representations.}
    \label{fig:framework-abla}
    \lxtailvs
    \vspace{-1em} 
    
\end{figure*}

We propose the Region-Aware Encoder (RAE), as shown in~\cref{fig:mainfig}. The key motivation is to use learnable parameters to adaptively balance the regional and background information. RAE employs a layer-wise coordination paradigm, uses self-attention to refine the regional feature and uses cross-attention to gather background information. 
Specifically, the image background is first encoded into contextual vision tokens using CE; then, for the ROI, it is first encoded using self-attention, followed by refinement through a cross-attention module. The query comes from the RAE, while the key and value come from the hidden states of the corresponding CE layer. This process can be formulated as:
\begin{equation}
\begin{aligned}
h_{\mathrm{RAE}}^{i+1} &= h_{\mathrm{RAE}}^{i} + \alpha \cdot \operatorname{xattn}\!\left(h_{\mathrm{RAE}}^{i},\, h_{\mathrm{CE}}^{i},\, h_{\mathrm{CE}}^{i}\right),\\
h_{\mathrm{CE}}^{i+1}  &= h_{\mathrm{CE}}^{i} + \operatorname{attn}\!\left(h_{\mathrm{CE}}^{i},\, h_{\mathrm{CE}}^{i},\, h_{\mathrm{CE}}^{i}\right),
\end{aligned}
\end{equation}
where $h_{RAE}^{i}$ is RAE's hidden state on layer $i$, $\alpha$ is a learnable weight parameter initialized to zero in the beginning, and $xattn$ is an operator that takes arguments in the order (q, k, v). 
To avoid training a vision encoder from scratch and reduce parameter count, self-attention modules in CE and RAE share weights across corresponding layers.
Beyond this, the layer-wise coordination capitalizes on the hierarchical features of ViTs, where shallow layers hold low-level details and deep layers contain high-level semantics~\cite{he2018layer-wise}. 
Compared with the encoder-decoder-like methods~\cite{lian2025dam} in~\cref{fig:framework-abla}(c), which only use the final-layer hidden states, RAE can capture more detailed cues from early layers, such as material and texture.

During training, since both the RAE and CE output visual tokens, we initially aligned them with the LLM via the same projector. However, we observed that training on region-level data would degrade CE performance, rendering RegRet unable to perform global-level retrieval using CE features. To address this issue, we shifted our perspective: the RAE maintains an independent semantic space, so we use a separate projector to align the RAE with the LLM and freeze the CE parameters. This simple yet effective decoupling design allows us to fully optimize the RAE without compromising the model's original representation capability. As a result, RegRet can retain its global retrieval performance.

\subsubsection{LMM For Multimodal Embedding}
Similar to prior work~\cite{jia2021scaling,liu2025lamra}, we add a special token \texttt{[EMB]} to the end of the instruction prompt (e.g., \texttt{"<image> Summarize the above image and sentence into one word: [EMB]"}). We use the last hidden state after the \texttt{[EMB]} as the embedding. The special token functions as a learnable query to summarize the information in the former sentence into an embedding.

\subsection{Training Pipeline}
\label{sec:methods-train-pipe}

\textbf{Stage-I: RAE Pretraining.} This stage serves to teach RAE to balance regional cues with background context and to enhance its fine‑grained understanding. We adopt the detailed localized captioning task~\cite{lian2025dam} as the pretext task, compelling the model to combine the background context and generate attribute-rich regional descriptions under the 
next-token prediction paradigm. We keep the RAE and its connector trainable and freeze the language backbone. Given the token sequence length $T$, token $x_i$, and RAE's trainable parameter $\theta$, the loss can be formulated as:
\begin{equation}
\mathcal{L}_{rae} = -\frac{1}{T} \sum_{t=1}^{T} \log P(x_t \mid x_{<t}; \theta).
\end{equation}

\par \noindent \textbf{Stage-II: Pure-Text Contrastive Learning.} In this stage, we perform contrastive learning with InfoNCE~\cite{oord2018representation} loss to efficiently transfer the LLM's language ability into embedding capabilities. We only use large-scale text-only pairs, as involving image–text pairs brings higher computational cost while yielding comparable performance~\cite{liu2025lamra}. After this phase, the model acquires an initial capacity for multimodal retrieval.

\par \noindent \textbf{Stage-III: Regional Contrastive Learning.} In the final stage, we conduct instruction tuning using a mixture of global-level and region-level contrastive pairs. Unlike previous methods~\cite{xie2025fgclip} which only align image regions with texts, we enable the construction of negative pairs between global and regional samples to enhance the retrieval performance at both levels. The language backbone is optimized with InfoNCE loss, which can be formulated as:

\begin{equation}
\mathcal{L}_{rcl} = -\frac{1}{N} \sum_{i=1}^N 
\log\!\left[
\frac{
\exp \left( 
\langle \Phi(q_i \otimes r_{q_i}), \Phi(c_i \otimes r_{c_i}) \rangle / \tau
\right)
}{
\sum_{j=1}^N 
\exp \left( 
\langle \Phi(q_i \otimes r_{q_i}), \Phi(c_j \otimes r_{c_j}) \rangle / \tau
\right)
}
\right]
\label{eq:rcl}
\end{equation}

where $\langle \cdot, \cdot \rangle$ denotes the inner product, $\tau$ is the temperature parameter, and $N$ is the batch size.

\section{REGMB: A Region-Level Multimodal Retrieval Benchmark}
\label{sec:bench}

\begin{figure*}[!t]
    \centering
    \includegraphics[width=\linewidth, keepaspectratio]{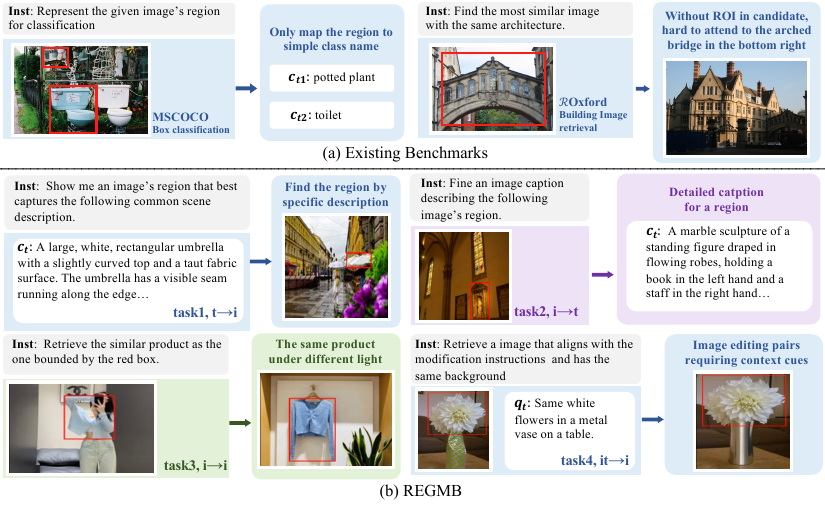}
    \lxcapvs
    \vspace*{-0.2em}
    \caption{Examples from REGMB and existing benchmarks (\S~\ref{sec:bench}). REGMB provides complex contrastive pairs and spans a wider range of modalities. $Inst$ denotes the task instruction, $c_t$ denotes text candidate, and $q_i$ denotes image query.}
    \label{fig:REGMB-case}
    \lxtailvs
    \vspace*{-1.8em}
\end{figure*}

In this section, we introduce REGMB\footnote{More detailed dataset statistics and preprocessing methods are provided in the appendices.}, a comprehensive \textbf{REG}ion-Level \textbf{M}ulti-modal Retrieval \textbf{B}enchmark. It fills two gaps in current studies: insufficient region-level training data and limited cross-modal evaluation tasks. Some data samples are shown in~\cref{fig:REGMB-case}. REGMB includes a total of 200k contrastive pairs for training and 25k for testing, collected from six diverse datasets and organized into two meta tasks that reflect different levels of retrieval granularity:

\par \noindent \textbf{Metatask 1: Region-Level Retrieval.} This metatask focuses on retrieving similar instances within the ROI. Although background information may aid interpretation, it is unnecessary for successful retrieval. This metatask includes three subtasks covering three retrieval settings: \textbf{(i) Task 1:} image-to-text retrieval, \textbf{(ii) Task 2:} text-to-image retrieval, and \textbf{(iii) Task 3:} image-to-image retrieval. 
For Tasks 1 and 2, we utilize public datasets with region-level annotations, including COCO~\cite{lin2014microsoft}, SAM~\cite{kirillov2023segment}, and FinHARD~\cite{xie2025fgclip}. For each region, we use the long, detailed captions generated by DAM~\cite{lian2025dam} and Qwen2.5-VL-72B to increase retrieval difficulty. When building Task 3, a unique challenge arises because \textit{existing public datasets lack cross-scene, region-level pairs of general objects}. For example, in Fashion200k, objects are typically captured at the same angle and under the same lighting conditions in a pure white scene. To overcome this limitation, we construct two data splits, \textit{XGoods} and \textit{XLife}, from a social media platform. We collect a set of image queries, retrieve daily life photos or products, and annotate the relevant pairs by human annotators. As illustrated in~\cref{fig:REGMB-case}(b), the blue shirt appears on the same person but in a different posture and under a different light source. Such variations significantly increase retrieval difficulty. All data are subjected to privacy filtering and manual region-level annotation to ensure high data quality.

\begin{table*}[!t]
\centering
\caption{\text{Comparison of methods on REGMB (\S~\ref{sec:analysis}).} RegRet-8B-zs, a zero-shot model without regional contrastive learning, already surpasses most baselines. The regional prompting strategies of baselines are in~\cref{sec:baselines}. IT2I denotes image-text-to-image retrieval. $^\dagger$ denotes models are trained with REGMB.}
\lxcapvs
\begin{adjustbox}{width=\textwidth, keepaspectratio}

\input{tabels/RegRet-bench}

\end{adjustbox}
\label{tab:RegRet-bench}
\lxtailvs
\vspace{-1em} 
\end{table*}

\begin{table}[t]
\caption{Generalization evaluation on existing region-level retrieval benchmarks (\S~\ref{sec:analysis}). The metric for $\mathcal{R}$Oxford-Hard, DeepFashion2, and ILIAS is mAP, recall@1, and mAP@50, respectively. $\mathcal{R}$Oxford only has ROIs in query.}
\lxcapvs
\label{tab:other-bench}
\centering
\setlength{\tabcolsep}{4pt}
\begin{adjustbox}{width=0.75\textwidth, keepaspectratio}
\begin{tabular}{l|c|c|cc|c}
\toprule
\multirow{2}{*}{Method} & \text{$\mathcal{R}$Oxford-Hard} & DeepFashion2 & \multicolumn{2}{c|}{ILIAS} & \multirow{2}{*}{Avg.} \\
 & I2I & I2I & I2I & T2I & \\
\midrule
VLM2VEC\cite{jiang2024vlm2vec}  & 24.4 & 6.2 & 36.1 & 27.1 & 23.5 \\
MMEMBED\cite{lin2024mmembed}  & 36.9 & 12.7 & 42.7 & 29.3 & 30.4\\
LamRA\cite{liu2025lamra} & 42.0 & 13.2 & 71.1 & \underline{62.6} & 47.2 \\
RzenEmbed\cite{jian2025rzenembed} &  33.9 & 9.5 & 42.4 & 38.6 & 31.1 \\
mmE5\cite{chen2025mmE5}  & 36.9 & 11.1 & 50.1 & 39.0 & 34.2 \\
\midrule
RegRet-8B-zs & \underline{42.8} & \underline{15.8} & \underline{75.7} & 59.1 & \underline{48.3} \\
RegRet-8B & \textbf{45.0} & \textbf{20.2} & \textbf{86.4} & \textbf{72.6} & 56.1\\
\bottomrule
\end{tabular}
\end{adjustbox}
\lxtailvs
\end{table}

\input{tabels/mbeir}

\par \noindent \textbf{Metatask2: Context-Level Retrieval.} This setting includes \textbf{Task 4}, image-text-to-image retrieval, where background context is critical for accurate retrieval. As illustrated in~\cref{fig:framework-abla}(b), multiple candidates might contain the flower, but only those with a metal vase in the background are considered correct. We leverage datasets such as \text{VisMin}~\cite{vismin2024} and \text{ImgDiff}~\cite{jiao2024imgdiffcontrastivedatasynthesis}, which naturally include region-level image pairs generated via image editing. 
Furthermore, we sample a subset from \text{XGoods} and automatically annotate it with Qwen2.5-VL-72B as complementary. 

\begin{table*}[t]
\centering
\caption{Ablations on the ViT architectures shown in~\cref{fig:framework-abla} (\S~\ref{sec:abla-vit}). All models are trained on REGMB and M‑BEIR based on the RegRet-8B.}
\lxcapvs
\label{tab:arch-abla}
\adjustbox{max width=\linewidth}{
\setlength{\tabcolsep}{3pt}
\begin{tabular}{l|ccc|ccccc|c}
    \toprule
    \text{Architecture} & xattn & \text{separate proj.} & layer-wise & Task1 & Task2 & Task3 & Task4 & REGMB Avg. & M-BEIR Avg.
 \\
    \midrule
    \text{Auxiliary Image} & \ding{55} & \ding{55} & \ding{55} & 69.8 & 82.7 & 92.6 & \textbf{86.7} &  83.4 & 57.1 \\  
    \text{RAE-sharep} & \ding{51} & \ding{55} & \ding{55} & 77.2 & 83.5 & 51.7 & 52.2 & 66.0 & 27.2 \\  
    \text{RAE-encdec} & \ding{51} & \ding{51} & \ding{55} & 76.6 & 83.8 & 89.4 & 81.3 & 82.6 & 57.1 \\
    \midrule
    \text{RAE} & \ding{51} & \ding{51} & \ding{51} &  \textbf{79.0} & \textbf{86.9} & \textbf{92.9} & \text{86.0} & \textbf{86.2} & \textbf{57.3}
 \\
    \bottomrule
\end{tabular}
}
\vspace{-0.8em} 
\end{table*}

\begin{table*}[t]
\caption{Ablations on the training data (\S~\ref{sec:abla-traindata}). We also trained LamRA from scratch with its original dataset and REGMB to show the effectiveness of our data. }
\lxcapvs
\label{tab:data-abla}
\centering
\adjustbox{max width=0.95\linewidth}{
\begin{tabular}{lc|ccc|cccccc}
\toprule
Model &Size & Language & Image-level & Regional & Task1 & Task2 & Task3 & Task4 & Avg.
 \\
\midrule
\multirow{3}{*}{RegRet} &  & \ding{51} &  & & 61.3 & 65.4 & 74.5 & 71.1 &  68.4 \\ 
 & \textit{3B} & \ding{51}  & \ding{51} &  &  65.1 & 62.4 & 64.6 & 80.1 &  69.4  \\
 & &  \ding{51} & \ding{51}  & \ding{51} &  \textbf{71.0} & \textbf{77.4} & \textbf{90.0} & \textbf{80.8} & \textbf{79.9} \\
\midrule
LamRA & \textit{8B} & \ding{51} & \ding{51} & \ding{51} & 64.0 & 79.2 & 89.0 & \textbf{87.8} &  80.9 \\
RegRet & \textit{8B} & \ding{51} & \ding{51} & \ding{51} & \textbf{79.0} & \textbf{86.9} & \textbf{92.9} & 86.0 & \textbf{86.2}\\
\bottomrule
\end{tabular}
}
\vspace*{-1em}
\end{table*}

\vspace{-0.4em}
\section{Experiments}
\vspace{-0.4em}
\subsection{Implementation Details}
We adopt Qwen2-VL~\cite{wang2024qwen2} models with 3B and 8B parameters as the backbone and train them using the three-stage strategy outlined in~\cref{sec:methods-train-pipe}. In Stage 1, we pretrain the Region-Aware Encoder using 800k region-level image-text pairs from the DAM~\cite{lian2025dam} and PAM~\cite{lin2025pam} datasets. During this stage, we only update the parameters of the RAE’s cross-attention layers and its projector. In stage 2, we used the datasets of Natural Language Inference (NLI)~\cite{gao2021simcse}, HotpotQA~\cite{yang2018hotpotqa}, and MSMARCO~\cite{MARCO}, totaling 780k text pairs.
In stage 3, we further finetune RegRet on a mixed dataset comprising 1.8M image-level pairs from M-BEIR~\cite{wei2023uniir} and 200k region-level pairs from REGMB. More training details are available in the supplementary material.

\subsection{Baselines}
\label{sec:baselines}
We evaluate our model against three categories of baselines: (i) LLM-based methods, such as MME5~\cite{chen2025mmE5}, VLM2VEC~\cite{jiang2024vlm2vec}, MMEMBED~\cite{lin2024mmembed}, LamRA~\cite{liu2025lamra}, and RzenEmbed~\cite{jian2025rzenembed}; (ii) CLIP-based methods, including SigLIP2~\cite{tschannen2025siglip}, BLIP2~\cite{li2023blip}, and UniIR~\cite{wei2023uniir};  and (iii) fine-grained region-level models, such as FG-CLIP~\cite{xie2025fgclip}, FineCLIP~\cite{lin2023fine}, and DreamLIP~\cite{zheng2024dreamlip}. To evaluate the model's zero-shot ability and demonstrate the effectiveness of the architecture design, we further provide \textbf{RegRet-8B-zs}, which is trained only on global-level data (\ie, M-BEIR) and not fine-tuned on a region-level dataset. 
To identify the optimal LMM regional  prompting strategy, we investigate multiple candidates. According to the results in~\cref{tab:prompt-abla}, we select the top-performing strategy per task: auxiliary images for I2T and T2I tasks, cropping for I2I task, and visual markers for IT2I task. For FG-CLIP, which supports ROIAlign, we use its native strategy.

\subsection{Evaluation Benchmarks}
\label{sec:eval-bench}
We evaluate RegRet at both regional and global levels. For regional retrieval, we primarily use REGMB, reporting Recall@5 for most tasks and Recall@1 for vismin and imgdiff to avoid metric saturation. To further assess its generalization, we include existing benchmarks like $\mathcal{R}$Oxford Hard split~\cite{radenovic2018revisitingoxford}, DeepFashion2~\cite{DeepFashion2}, and ILIAS~\cite{ilias}. Notably, $\mathcal{R}$Oxford does not have ROIs in its candidate pool. For global-level evaluation, we adopt the M-BEIR~\cite{wei2023uniir} benchmark.

\subsection{Analysis}
\label{sec:analysis}
Our main results are shown in~\cref{tab:RegRet-bench} and ~\cref{tab:other-bench}. For both REGMB and public datasets, our method outperforms strong baselines. Several key observations can be drawn from the results: 

\par \noindent \textbf{(1) The effectiveness of model design.} In the zero-shot setting, RegRet-8B-zs outperforms all the baselines on REGMB and public benchmarks. It exceeds the average performance of LMMs that employ regional prompting strategies by 14.3\% and 15.0\%, respectively. It is even on par with the fine-tuned leading LMM in~\cref{tab:data-abla}. For $\mathcal{R}$Oxford, which does not have ROIs in candidates, our method can still perform well. This is mostly due to the RAE's balance between regional and background information.

\par \noindent \textbf{(2) The effectiveness of training data and strategy.} After training on REGMB, the 3B model already surpasses some 11B models (e.g., mmE5), and RegRet-8B achieves an extra average improvement of 7.4\% compared to the zero-shot version. The performance gain generalizes to unseen public benchmarks, yielding an average 22.8\% improvement, demonstrating the effectiveness of our data and training strategy.

\par \noindent \textbf{(3) RegRet can preserve or even enhance global-level retrieval performance.} As reported in~\cref{tab:mbeir}, we observe performance gains compared to LamRA on the majority of splits in M-BEIR. Since novel models like mmE5 and RzenEmbed have made many improvements to training data and the loss function, we argue that LamRA is the most reasonable baseline for examining global-level performance. The gains are particularly notable on IT2IT tasks like InfoSeek, where it increases by 1.2\%. This indicates that training RegRet on regional contrastive data can at least preserve global-level retrieval ability.

\par \noindent \textbf{(4) Inherent tension exists between local and global representation learning.} Models that perform well in Tasks 1–3 (\eg, SigLIP2) may underperform in Task 4, and vice versa (\eg, MME5). For Tasks 1-3, the model needs to exclude background noise, while the background information is crucial in Task 4. Models that focus on region-level features are less effective at global-level understanding. Therefore, RegRet also slightly underperforms some models on Task 4. Despite this, our method achieves a sweet spot in this trade-off by decoupling RAE and CE, enabling it to perform effectively across all tasks.

\subsection{Ablations}
In this section, we conduct ablations to further verify the effectiveness of each module. We choose LamRA as the LMM for comparison because it shares a similar training procedure and dataset with RegRet, without synthetic data (\eg, mmE5) and with an improved InfoNCE loss (\eg, RzenEmbed). Therefore, it ensures a fair validation of RegRet's performance.

\subsubsection{RAE Architecture}
\label{sec:abla-vit}
To validate the effectiveness of each design in RAE, we compare it with different architectures, as shown in~\cref{tab:arch-abla}. RAE-sharep use a single projector using the last-layer's hidden states. RAE-encdec adds separate projectors, and RAE further employs the layer-wise coordination. All the variants are based on RegRet-8B and undergo the same training process. We can observe that using separate projectors can prevent interference between local and global information. RAE-encdec avoids the giant performance degradation compared to RAE-sharep. Moreover, although the auxiliary image strategy is comparable to the RAE-encdec on REGMB, it still underperforms on Tasks 1 and 2, as analyzed in~\cref{sec:case-study}. 
RAE incorporates layer-wise coordination and further enhances region-level performance, as it can align CE and RAE features across different semantic levels.

\subsubsection{Training Data.}
\label{sec:abla-traindata}
~\cref{tab:data-abla} presents ablations on different training data sources. We compare the use of pure-text data from Stage-II, image-level data from M-BEIR, and regional-level data from REGMB. 
We observe a notable boost in regional retrieval accuracy after integrating region-level contrastive pairs. Training LamRA with REGMB also improves its performance (67.7\%$\longrightarrow$80.9\%). This reveals the effectiveness of our data.

\subsubsection{Regional Prompt Strategy}
\label{sec:abla-prompt}

\begin{wraptable}{r}{0.63\linewidth}
\vspace{-3em}
\caption{
Zero-shot performance of RegRet and LMMs with different prompting strategies on REGMB (\S~\ref{sec:abla-prompt}).
 Only using prompts to adapt LMMs models to region-level retrieval proves suboptimal.
}
\label{tab:prompt-abla}
\centering
\adjustbox{max width=\linewidth}{
\begin{tabular}{lccccc}
    \toprule
    Prompt & Task1 & Task2 & Task3 & Task4 & Avg. \\
    \midrule
    Crop              & 55.3 & 28.3 & 79.2 & 59.2 & 55.9 \\
    Visual mark          & 39.8 & 40.5 & 55.7 & 77.1 & 55.9 \\
    Auxiliary image   & 48.1 & 52.5 & 70.4 & 75.2 & 63.1 \\
    \midrule
    RegRet-8B-zs       & \textbf{69.9} & \textbf{76.7} & \textbf{85.7} & \textbf{81.6} & \textbf{78.8} \\
    \bottomrule
\end{tabular}
}
\vspace{-2em}
\end{wraptable}

To demonstrate that directly transferring regional prompting strategies to the retrieval task underperforms RAE, we equip LamRA with multiple strategies and compare it with RegRet-8B-zs. The results are reported in~\cref{tab:prompt-abla}. 
We evaluate three prompt methods as described in~\cref{sec:methods-model-design}, including visual mark, crop, and auxiliary image. 
Results show that cropping and visual mark are close in performance, while the auxiliary image yields substantial improvements.
However, even with the most effective strategy, LamRA still falls short of RegRet, suggesting that simply adding or cropping an image is insufficient to achieve the same performance gains.

\begin{figure}[t]
\centering
\includegraphics[keepaspectratio,width=\textwidth]{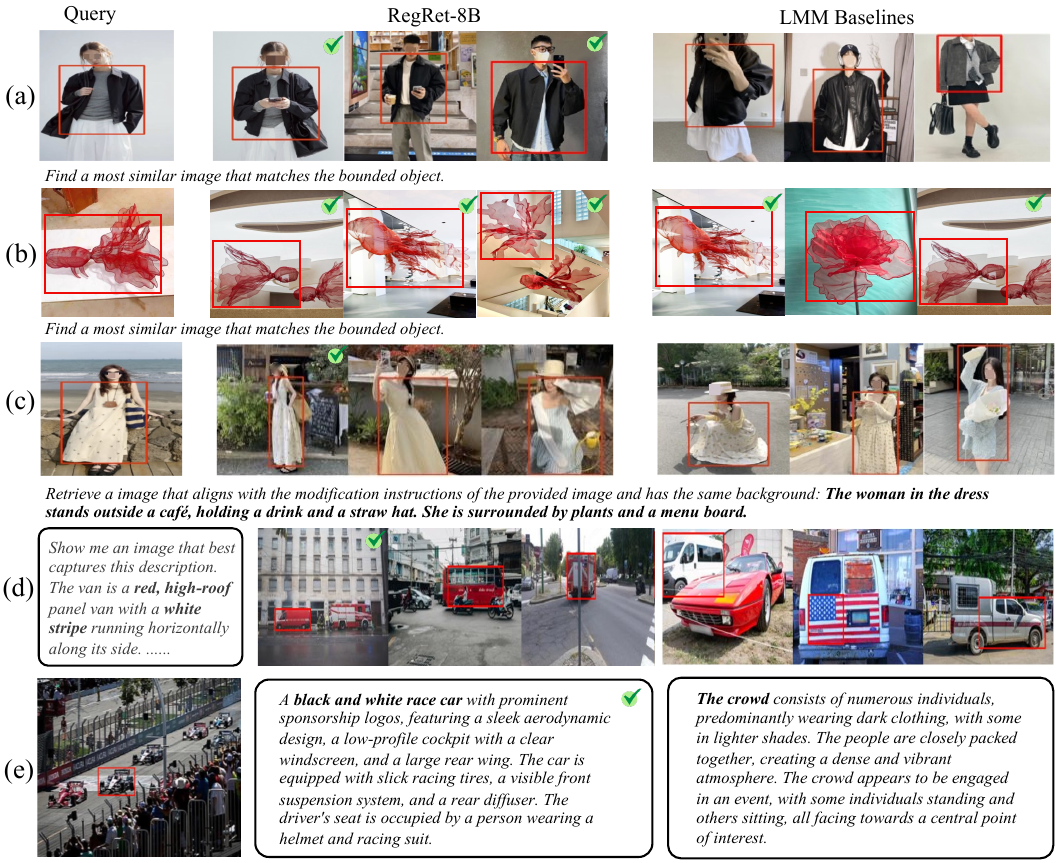}
\vspace{-1em}
\caption{Queries and top three candidates from REGMB (\S~\ref{sec:case-study}). All the results are retrieved from the around 5k candidate pool of each subtask. The positive results are labelled with green checkmarks. }
\vspace{-2em}
\label{fig:case-t34}
\end{figure}

\vspace{-0.4em}
\section{Case Study}
\vspace{-0.4em}
\label{sec:case-study}
In this section, we conduct a detailed case study to illustrate why RegRet outperforms existing LMMs across various regional prompting strategies. Retrieval samples from RegRet-8B and LamRA are listed in~\cref{fig:case-t34}. We can make the following observations: 
\textbf{(1) RegRet can incorporate background context. }In~\cref{fig:case-t34}(c), it selects the image showing a woman in front of a café. Although other candidates wear similar clothing, they are rejected because they do not meet the user's background requirements. \textbf{(2) RegRet can discard background distractors, while auxiliary image may introduce it.} As shown in~\cref{fig:case-t34}(e), RegRet only focuses on the racecar within the bounding box. However, when using the baseline LMM with auxiliary images, it incorrectly matches the image to ``crowd'', which dominates the background. 
\textbf{(3) Cropping may lose background for accurately interpreting ROI.} In~\cref{fig:case-t34}(b), when LamRA uses cropped regions to retrieve, it mistakenly interprets a flower for the golden fish decoration. If it is fed the green background, the model can understand that the candidate is a handicraft rather than a building decorative element.
\textbf{(4) RegRet has fine-grained visual understanding ability.} In~\cref{fig:case-t34}(a), while the baseline returns leather black jackets, RegRet captures the material feature and finds the correct cloth ones. This ability arises from the RAE pretraining stage using detailed captions.

\vspace{-0.4em}
\section{Conclusion}
\vspace{-0.4em}

We present RegRet, a novel framework designed for diverse region-level multimodal retrieval tasks. Equipped with the Region-Aware Encoder, it significantly enhances region-level representation. We further build the REGMB benchmark to enable the training and evaluation of regional retrieval models.
Compared to previous approaches, RegRet sets a new state-of-the-art in performance. Future work will explore extending RegRet’s ability to visual document retrieval and integrating it into retrieval-augmented generation systems.

\section*{Acknowledgements}
This work was supported in part by the National Nature Science Foundation of China (Grant No: 62273303, 62273302), in part by Yongjiang Talent Introduction Programme (2022A-240-G).

%
%

\bibliographystyle{splncs04}
\bibliography{main}

\include{supp.tex}

\end{document}

%% file: tabels/RegRet-bench.tex
\begin{tabular}{lcccccccccccc}
\toprule
\multirow{3}{*}{Methods} &
\multirow{3}{*}{Size} &
  \multicolumn{2}{c}{\text{Task 1 (T2I)}} &
  \multicolumn{2}{c}{\text{Task 2 (I2T)}} &
  \multicolumn{2}{c}{\text{Task 3 (I2I)}} &
  \multicolumn{3}{c}{\text{Task 4 (IT2I)}} &
\multirow{3}{*}{Avg.} \\
\cmidrule(lr){3-4}
\cmidrule(lr){5-6}
\cmidrule(lr){7-8}
\cmidrule(lr){9-11}
& & sam & coyo & sam & coyo & xlife & xgoods & vismin & imgdiff & xgoods & \\
\cmidrule(lr){3-4}
\cmidrule(lr){5-6}
\cmidrule(lr){7-8}
\cmidrule(lr){9-11}
& & R@5 & R@5 & R@5 & R@5 & R@5 & R@5 & R@1 & R@1 & R@5 & \\
\midrule
\textbf{\textit{CLIP-based}}  & & & & & &  & & & & & \\
FG-CLIP\cite{xie2025fgclip} & \textit{0.2B} & 48.5 & 58.7 & 48.9 & 57.4 & 50.0 & 60.7 & 53.6 & \textbf{82.2} & 66.6 & 58.5\\
SigLIP2\cite{tschannen2025siglip} & \textit{0.4B} & 46.3 & 74.2 & 55.9 & 74.7 & 76.9 & 79.0 & 30.5 & 51.6 & 76.7 & 62.9 \\
DreamLIP\cite{zheng2024dreamlip} & \textit{0.4B} & 37.9 & 63.3 & 45.9 & 71.3 & 49.1 & 49.3 & 55.7 & 54.9 & 69.6 & 55.2\\
FineCLIP\cite{jing2024fineclip} & \textit{0.4B} & 39.4 & 72.9 & 44.2 & 71.4 & 60.5 & 67.7 & 57.2 & 28.9 & 23.7 & 51.8\\
EVA-CLIP\cite{sun2024eva} & \textit{8B} & 48.5 & 58.7 & 48.9 & 57.4 & 50.0 & 60.7 & 53.6 & \textbf{82.2} & 66.6 & 58.5\\
\midrule
\textbf{\textit{LMM-based}} & & & & & &  & & & & & \\
MM-EMBED\cite{lin2024mmembed} & \textit{8B} & 42.5 & 46.4 & 35.2 & 32.1 & 72.5 & 80.5 & 82.9 & 72.5 & 87.0 & 61.3\\
VLM2VEC\cite{jiang2024vlm2vec} & \textit{8B} & 39.8 & 49.9 & 37.3 & 48.5 & 58.7 & 68.0 & 79.3 & 72.2 & 90.1 & 60.4 \\

LamRA\cite{liu2025lamra} & \textit{8B} & 41.3 & 55.0 & 46.7 & 58.3 & 75.0 & 83.3 & 82.6 & 62.8 & 86.0 & 65.7 \\
RzenEmbed\cite{jian2025rzenembed} & \textit{8B} & 41.9 & 56.6 & 59.1 & 73.6 & 50.0 & 68.3 & \textbf{86.9} & 72.5 & \textbf{98.2} & 67.5\\
mmE5\cite{chen2025mmE5} & \textit{11B} & 50.4 & 55.2 & 42.2 & 51.4 & 75.8 & 82.0 & 80.9 & 75.3 & \underline{94.4} & 67.5\\
\midrule
\textbf{\textit{Ours}}  & & & & & &  & & & & & \\
RegRet-3B\text{$^\dagger$} & \textit{3B} & 58.1 & \underline{83.9} & 70.6 & \underline{84.3} & \underline{88.3} & \underline{91.7} & 83.1 & 78.0 & 81.3 & \underline{79.9} \\
RegRet-8B-zs & \textit{8B} & \underline{59.5} & 80.3 & \underline{72.4} & 80.9 & 85.1 & 86.3 & 81.1 & 70.8 & 92.8 & 78.8 \\
RegRet-8B\text{$^\dagger$} & \textit{8B} & \textbf{69.1} & \textbf{88.9} & \textbf{86.5} & \textbf{87.3} & \textbf{92.3} & \textbf{93.6} & \underline{83.4} & \underline{81.1} & 93.6 & \textbf{86.2} \\
\bottomrule
\end{tabular}

%% file: tabels/mbeir.tex
\begin{table*}[!t]
\centering
\caption{Performance on the global-level benchmark, M-BEIR (\S~\ref{sec:analysis}). Although specifically optimized for region-level tasks, RegRet maintains its global-level retrieval capibility and even achieves better results. $^\dagger$ denotes models trained with M-BEIR.} 
\lxcapvs
\resizebox{\linewidth}{!}{

\begin{tabular}{lc@{\hspace{0.1cm}}c@{\hspace{0.1cm}}c@{\hspace{0.1cm}}c@{\hspace{0.1cm}}c@{\hspace{0.1cm}}c@{\hspace{0.1cm}}c@{\hspace{0.1cm}}c@{\hspace{0.1cm}}c@{\hspace{0.1cm}}c@{\hspace{0.1cm}}c@{\hspace{0.1cm}}c@{\hspace{0.1cm}}c@{\hspace{0.1cm}}}
\toprule
 & \multicolumn{2}{c}{T2I} & \multicolumn{2}{c}{T2IT} & \multicolumn{2}{c}{I2T} & {I2I} & \multicolumn{1}{c}{IT2T} & \multicolumn{2}{c}{IT2I} & \multicolumn{2}{c}{IT2IT} & \\
 \cmidrule(r){2-3} \cmidrule(r){4-5} \cmidrule(r){6-7} \cmidrule(r){8-8} \cmidrule(r){9-9} \cmidrule(r){10-11} \cmidrule(r){12-13} 
 Methods & COCO & F200K & EDIS & WebQA & COCO & F200K & NIGHTS & InfoSeek & F200IQ & CIRR & OVEN & InfoSeek & Avg. \\
\cmidrule(r){2-3} \cmidrule(r){4-5} \cmidrule(r){6-7} \cmidrule(r){8-8} \cmidrule(r){9-9} \cmidrule(r){10-11} \cmidrule(r){12-13} 
& R@5 & R@10 & R@5 & R@5 & R@5 & R@10 & R@5 & R@5 & R@10 & R@5 & R@5 & R@5 & \\
\midrule
SigLIP\cite{zhai2023sigmoid}  & 75.7 & \textbf{36.5} & 27.0 & 43.5 & 88.2  & \textbf{34.2} & 28.9 & 25.1 & 14.4 & 22.7 & 41.7 & 27.4 & 38.8  \\
BLIP2\cite{li2023blip} & 63.8 & 14.0 & 26.9 & 24.5 & 80.0  & 14.2 & 25.4 & 5.5 & 4.4 & 11.8 & 27.3 & 15.8 & 26.1  \\
Qwen2-VL-7B\cite{wang2024qwen2} & 55.1 & 5.0 & 26.2 & 9.4 & 46.6 & 4.0 & 21.3 & 22.5 & 4.3 & 16.3 & 43.6 & 36.2 & 22.3 \\
\midrule
$\text{UniIR-BLIP}_{\text{FF}}\text{$\dagger$}$\cite{wei2023uniir} & 79.7 & 26.1 & 50.9 & 79.8 & 89.9 & 28.9 & \textbf{33.0} & 22.4 & 29.2 & 52.2 & 55.8 & 33.0 & 48.4  \\
$\text{UniIR-CLIP}_{\text{SF}}\text{$\dagger$}$\cite{wei2023uniir} & 81.1 & 18.0 & 59.4 & 78.7 & \textbf{92.3} & 18.3 & 32.0 & 27.9 & 24.4 & 44.6 & 67.6 & 48.9 & 49.4  \\
LamRA\text{$\dagger$}\cite{liu2025lamra} & \underline{81.5} & 28.7 & \underline{62.6} & \underline{81.2} & \underline{90.6} & 30.4 & \underline{32.1} & \underline{52.1} & \textbf{33.2} & \underline{53.1} & \underline{76.2} & \underline{63.3} & \underline{57.1} \\
RegRet-8B\text{$\dagger$} & \textbf{81.6} & \underline{28.9} & \textbf{62.7} & \textbf{81.8} & 90.3 & \underline{30.6} & 31.6 & \textbf{52.7} & \textbf{33.2} & \textbf{53.3} & \textbf{76.4} & \textbf{64.5} & \textbf{57.3}\\
\bottomrule
\end{tabular}
}
\lxtailvs
\vspace{-1.5em}

\label{tab:mbeir}
\end{table*}

%% file: supp.tex
\clearpage
\setcounter{page}{1}

\appendix
\section*{Appendices}
\label{sec:appendices}

\section{Incorporating RoIs into Candidates}

A key observation in region-level retrieval is the necessity of providing an explicit indicator, such as a RoI, to guide the model's attention to the relevant area. Without such guidance, the model often relies on global visual similarity, leading to suboptimal retrieval performance.

For instance, in ~\cref{fig:case-t34}(d), the negative candidates retrieved by LamRA share a dominant visual pattern: red vehicles with white stripes occupying most of the image. This suggests that the model is aligning the text with the full image rather than focusing on the specific region that better matches the query. 
This is because the model has to represent the full image with a single embedding when there are no extra modules to represent the RoI. The global-level embedding inevitably overlooks some local features and only utilizes the dominant visual cues as candidates. In the LamRA's semantic space, the true positive candidate image is closer to \emph{``Two cars in front of a building''}, instead of \emph{``A red van with white strips''}. 
Another example is in~\cref{fig:case-t34}(e). The top-ranked negative candidate shares not only a similar outfit with the query but also a comparable background, including a horizon line located at one-third of the image height. However, this background similarity is misleading. The retrieval task only requires consistency in the foreground region, while the background, which should ideally have a café and a menu board, can differ from the query.
Therefore, the standard approach to address these challenges is to include RoIs in both query and candidate inputs~\cite{zhang2018visual,faysse2025colpaliefficientdocumentretrieval}. This allows the model to focus on the relevant region. At the same time, background context can be downweighted, reducing the risk of interference from irrelevant visual features.


\section{Inference Time Analysis}

\begin{figure}[htbp]
\centering
\includegraphics[keepaspectratio]{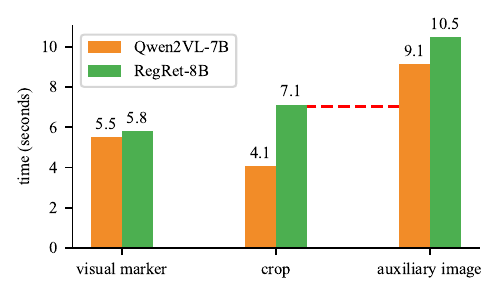}
\caption{Comparison of inference time with Qwen2VL-7B. The average time for generating a batch of embeddings on the I2I task of REGMB is reported. Compared with auxiliary image, which achieves the best average performance on REGMB, RegRet is 22\% faster that baseline models.}
\label{fig:infer-time}
\end{figure}

To quantitatively assess the additional inference time introduced by incorporating RAE, we compare its speed with the base model, Qwen2VL-7B, under different prompting strategies. The evaluation is conducted on samples with varying image sizes and target region scales. We report the average inference time per batch in~\cref{fig:infer-time}, using a batch size of 96.
Under the visual marker strategy, both RegRet and Qwen2VL-7B process the full image only once, resulting in comparable inference times. Under the crop strategy, RegRet is slower than Qwen. This slowdown occurs because RegRet encodes the entire image using the CE and also processes the RoI with RAE. In comparison, Qwen2VL-7B encodes only the RoI with CE, which is typically a small region, with considerably fewer visual tokens under Qwen's dynamic resolution paradigm. Meanwhile, given that RegRet achieves more than 20\% performance improvement over Qwen-based baselines using the crop strategy, the additional inference cost is acceptable. 

As to the auxiliary image strategy, this performance gap is narrowed. In this setting, both models encode two images: Qwen2VL-7B uses CE for both, while RegRet uses CE for the full image and RAE for the RoI. 
In practice, under the scenarios, such as task 4 on REGMB, where Qwen needs the auxiliary image strategy to fully boost their performance, RegRet is even faster (9.1$\longrightarrow$7.1), as illustrated in~\cref{fig:infer-time} with the red dashed line.


\section{Training Details}
\label{sec:app-train}
\subsection{Training Parameters}
In the first stage, we use the batch size 64 and the learning rate of $1 \times 10^{-4}$. Only RAE's cross-attention modules and the projector are trainable.
In the second stage, the batch size is 1104 and the learning rate is $1.1 \times 10^{-4}$. We only update the language model backbone's parameters for one epoch using pure-text contrastive pairs. 
In the final stage, we fine-tune the language model backbone for two epochs on multimodal contrastive pairs, with a learning rate of $2 \times 10^{-4}$ and a batch size of 1104. 
To save the GPU memory for a larger batch size, which is crucial in contrastive learning, we use LoRA to fine-tune the LLM. The LoRA rank is set to be 64 and 128 for stage two and stage three, respectively. All three stages can be completed within 20 hours on 24 NVIDIA H20 GPUs. 

\subsection{Training Strategy Ablations}

\begin{table}[htbp]
\centering
\caption{Ablations on the training strategy in Stage-III using 3B models, evaluated on REGMB.
}
\label{tab:train-strategy}
\adjustbox{max width=\linewidth}{
\setlength{\tabcolsep}{3pt}
\begin{tabular}{lccccc}
\toprule
\textbf{Strategy} & \textbf{Task1} & \textbf{Task2} & \textbf{Task3} & \textbf{Task4} & \textbf{Avg.} \\
\midrule
crop  & 68.2 & 78.4 & 89.3 & 77.9 &  78.4 \\
concat  & 68.3 & 77.3 & 88.5 & 71.4 &  75.8  \\
random  & 67.8 & 78.9 & 89.2 & 73.0 &  76.7 \\
mixed visual prompts  & 71.0 & 77.4 & 90.0 & 80.8 &  79.9 \\
\bottomrule
\end{tabular}
}
\end{table}

During the training stage three, we develop the \emph{mixed visual prompting strategy} to balance contextual and regional information. In particular, for tasks 1 to 3 in the REGMB training split, which rely more on local details, we provide only RAE tokens to the LLM. For task 4, which requires contextual understanding, we concatenate CE tokens with RAE tokens. This design allows the model to leverage CE tokens for context without injecting excessive background signals into RAE, thereby preserving fine-grained representation quality. 

We compare the mixed visual prompting strategy with some approaches in~\ref{tab:train-strategy}. This strategy proves to be able to provide consistent performance improvements across both meta tasks. As shown in Table~\ref{tab:train-strategy}, we compare several alternative training strategies. The ``crop'' setting uses only the RAE tokens for both meta tasks; the ``concat'' setting uses the concatenation of context encoder (CE) and RAE tokens as visual input; and the ``random'' strategy randomly selects either ``crop'' or ``concat'' with a 50\% probability for each meta task. We observe that our strategy achieves the best performance. This is mainly because of the trade-off between the two metatasks. Specifically, metatask 2 requires contextual information for comprehensive representation, while metatask 1 demands a highly localized, fine-grained understanding. Excessive context may thus act as a distraction in the latter. Therefore, we provide the CE tokens when training on metatask2, so that the model can access the global representations. Meanwhile, we use only RAE tokens on metatask 1 to prevent introducing background noise. 

\section{Public Evaluation Benchmarks}
In this section, we introduce the public datasets used in our experiments. As publicly available datasets with ROI annotations are relatively scarce, we carefully select three widely used, representative benchmarks with standard retrieval evaluation protocols for performance testing. Since most of these benchmarks are designed for image-to-image (I2I) retrieval, we apply a unified cropping strategy for all baseline models to ensure fair comparison, as discussed in~\cref{sec:baselines}.
\par \noindent \textbf{$\mathcal{R}$Oxford.} This dataset covers building scenes in Oxford, and is split into easy and hard subsets, with ROIs provided only for query images. In this work, we report results on the hard subset. We follow the original evaluation protocol and use mean Average Precision (mAP) as the metric. The top-50 metric fails to reflect actual retrieval performance here, because each query is associated with a large number of junk samples.
\par \noindent \textbf{DeepFashion2.} DeepFashion2 is a multi-task fashion dataset. We adopt its consumer-to-in-shop retrieval task, which aims to retrieve the most similar clothing items from the in-shop gallery based on user-uploaded consumer photos. Since the official test set candidate pool is not publicly available, we follow common practice and run experiments on the validation set. We report average Recall@1, which measures whether the positive candidate for each query outranks all negative samples. This is a stricter metric than mAP for this task.
\par \noindent \textbf{ILIAS.} This benchmark supports region-level retrieval on general images, covering both I2I and image-to-text (I2T) tasks. The original dataset includes a 1M distractor pool to increase retrieval difficulty. As indexing 1M candidates is excessively time-consuming, we use the 4.72k core subset for evaluation, and follow the original paper to adopt mAP@50 as the metric.

\section{Dataset Details}
\label{sec:app-dataset}

This section details the construction of REGMB. The benchmark is aggregated from a total of six data sources. These include four publicly available datasets-SAM, COYO, ImgDiff, and Vismin-and two proprietary datasets: XGoods and XLife, which we collected from a social media platform to cover a broader range of fused-modal scenarios. Using the bounding box annotations from these datasets, we created regional query-candidate pairs. To guarantee that the benchmark is sufficiently challenging and discriminative, each query is paired with one to six hard negatives across all tasks.

\par \noindent \textbf{XGoods and XLife.} Recognizing the limited availability of public datasets for region-level image-to-image retrieval, we collected XGoods from an social media platform. Unlike datasets such as NIGHTS, which only contain the same object in similar backgrounds, XGoods offers more complex and realistic scenes. In this task, a commercial product image acts as the query, while a user-uploaded photo of the same item serves as the positive candidate. We use a fine-tuned YOLO v8 model to detect the key objects, and use the regional feature to retrieve similar images using SigLIP2. Then we employ human annotators to select the relevant samples as positive candidates, and keep the rest of them as hard negatives. We paired each query with six highly relevant products as hard negatives. For Task 3, we allocated 69k samples for training and 3.3k for evaluation. For Task 4, we also built a 3.4k data test split as a complementary dataset.

Similar to XGoods, XLife was also collected from social media content on the same platform. The main difference is its focus on images from users' daily lives rather than commercial product displays. The data was processed using the same procedure as XGoods, from which we carefully selected 4.2k query-candidate pairs for the final test set. Please refer to~\cref{fig:caption-prompt2} for the prompt used for generating descriptions.

\begin{figure}[htbp]
    \centering
    \begin{tcolorbox}[
        enhanced,
        colback=boxbg,
        colframe=boxframe,
        coltitle=white,
        colbacktitle=boxtitle,
        fonttitle=\bfseries\small,
        title={\quad Prompt},
        arc=6pt,
        boxrule=1.2pt,
        left=8pt, right=8pt, top=6pt, bottom=6pt,
        width=0.9\linewidth
    ]
    \ttfamily\small
You are a contrastive learning data annotator skilled at describing differences and relationships between images to generate query-candidate pairs. Given an image of a query object and a set of candidate images, for each candidate image, describe the environment where the query object appears, its relationship with surrounding objects, or changes in its own attributes. Examples:  

a. A woman wearing this top is sitting in a studio broadcasting.  

b. The same sofa is placed in a living room with a coffee table in front.  

c. A green bottle of toner with a small cat sitting to its left.  

d. A teddy bear of identical shape but brown instead of beige, placed among a group of dolls.  

\text{Description rules:}  

a. Keep descriptions concise, under 100 words.  

b. Include sufficient detail to clearly convey the specific context of the query object in the candidate image.  

c. Ensure uniqueness: the combination of the query and description should uniquely identify the corresponding candidate image.  

\text{Output format:  }

a. Each candidate image receives one description and a score (between 0 and 1) indicating the relevance and accuracy of the description. Assign 0 if the candidate is either too similar to the query or lacks distinctive contextual details making description difficult; assign 1 if the description is clear and highly discriminative.  
b. Descriptions must be in English.  

c. Strictly follow JSON format as shown below:  

{

"candidates": [{
  
    "0": "description0",
    
    "score": "0.7"
    
  },{
  
    "1": "description1",
    
    "score": "0.1"
    
  }]}
    \end{tcolorbox}
    \caption{The prompt for generating descriptions on Task 4, XGoods. }
    \label{fig:caption-prompt2}
\end{figure}

\par \noindent \textbf{SAM.} This dataset is used to build Tasks 1 and 2. We first converted the segmentation masks from SAM into bounding boxes by taking the maximum coordinates along each axis, and discard those too small regions. Each region was generated using Qwen2.5-VL-72B with auxiliary image, as shown in~\cref{fig:caption-prompt}. 
To ensure an adequate supply of hard negatives, we required that each candidate pool include at least two distinct regions. For both Task 1 and Task 2, this process yielded 30k samples for training and 1.5k for evaluation. We further manually checked the relevace of the true positive candidates in the evaluation dataset.

\begin{figure}[h]
    \centering
    \begin{tcolorbox}[
        enhanced,
        colback=boxbg,
        colframe=boxframe,
        coltitle=white,
        colbacktitle=boxtitle,
        fonttitle=\bfseries\small,
        title={\quad Prompt: Detailed Image Caption Generation},
        arc=6pt,
        boxrule=1.2pt,
        left=8pt, right=8pt, top=6pt, bottom=6pt,
        width=0.9\linewidth
    ]
    \ttfamily\small
    Describe the target region bounded by a red box in detail. Incorporate the background to ensure a property understanding of the region.
    \end{tcolorbox}
    \caption{The prompt for generating regional captions. }
    \label{fig:caption-prompt}
\end{figure}

\par \noindent \textbf{COYO.} From the COYO dataset, we utilized bounding boxes previously generated by FG-CLIP and subsequently re-captioned each region with DAM. Employing much longer captions allows for a richer alignment between the image region and its associated semantic concepts. These region-caption pairs naturally form the basis for text-to-image and image-to-text retrieval tasks, for which we prepared 1.5k evaluation samples each.

\par \noindent \textbf{ImgDiff.} As an image editing dataset, each sample in ImgDiff contains a source image, an editing instruction, an edited image, and a bounding box indicating the modified area. This structure lends itself naturally to an image-text-to-image retrieval task, where the query is a composite of the source image and the instruction, and the positive candidate contains the target region. When constructing the test split, we exclusively retained samples with two or more editing instructions to serve as hard negatives. This split comprises 21k training samples and 1.3k test samples.

\par \noindent \textbf{Vismin.} The structure of Vismin is analogous to that of ImgDiff; the primary difference is that Vismin contains real-world photographs, whereas ImgDiff's images are in the AI-generated style. We applied the same processing approach: the source image and edit instruction from the query, and the modified image serves as the positive candidate. From this dataset, we selected 46k samples for training and 1.3k queries for testing.



%% file: main.bib
@String(CVPR  = {IEEE Conf. Comput. Vis. Pattern Recog.})

@String(ICCV  = {Int. Conf. Comput. Vis.})

@String(ECCV  = {Eur. Conf. Comput. Vis.})

@String(ICML  = {Int. Conf. Mach. Learn.})

@String(CVPR  = {CVPR})

@String(ICCV  = {ICCV})

@String(ECCV  = {ECCV})

@String(ICML  = {ICML})

@String(icml = {Proceedings of the International Conference on Machine Learning})

@String(emnlp = {Proceedings of the Conference on Empirical Methods in Natural Language Processing})

@inproceedings{wei2023uniir,
  title={Uniir: Training and benchmarking universal multimodal information retrievers},
  author={Wei, Cong and Chen, Yang and Chen, Haonan and Hu, Hexiang and Zhang, Ge and Fu, Jie and Ritter, Alan and Chen, Wenhu},
  booktitle=eccv,
  year={2024}
}

@inproceedings{liu2025lamra,
  title={Lamra: Large multimodal model as your advanced retrieval assistant},
  author={Liu, Yikun and Zhang, Yajie and Cai, Jiayin and Jiang, Xiaolong and Hu, Yao and Yao, Jiangchao and Wang, Yanfeng and Xie, Weidi},
  booktitle={Proceedings of the Computer Vision and Pattern Recognition Conference},
  pages={4015--4025},
  year={2025}
}

@article{jiang2024e5v,
  title={E5-V: Universal Embeddings with Multimodal Large Language Models},
  author={Jiang, Ting and Song, Minghui and Zhang, Zihan and Huang, Haizhen and Deng, Weiwei and Sun, Feng and Zhang, Qi and Wang, Deqing and Zhuang, Fuzhen},
  journal={arXiv preprint arXiv:2407.12580},
  year={2024}
}

@misc{lin2024mmembed,
  title        = {MM-Embed: Universal Multimodal Retrieval with Multimodal LLMs},
  author       = {Sheng-Chieh Lin and Chankyu Lee and Mohammad Shoeybi and Jimmy Lin and Bryan Catanzaro and Wei Ping},
  year         = {2024},
  eprint       = {2411.02571},
  archivePrefix= {arXiv},
  primaryClass = {cs.CL},
  url          = {https://arxiv.org/abs/2411.02571}
}

@inproceedings{xiao2025flair,
  title={Flair: Vlm with fine-grained language-informed image representations},
  author={Xiao, Rui and Kim, Sanghwan and Georgescu, Mariana-Iuliana and Akata, Zeynep and Alaniz, Stephan},
  booktitle={Proceedings of the Computer Vision and Pattern Recognition Conference},
  pages={24884--24894},
  year={2025}
}

@inproceedings{zhong2022regionclip,
  title={Regionclip: Region-based language-image pretraining},
  author={Zhong, Yiwu and Yang, Jianwei and Zhang, Pengchuan and Li, Chunyuan and Codella, Noel and Li, Liunian Harold and Zhou, Luowei and Dai, Xiyang and Yuan, Lu and Li, Yin and others},
  booktitle={Proceedings of the IEEE/CVF conference on computer vision and pattern recognition},
  pages={16793--16803},
  year={2022}
}

@inproceedings{kirillov2023segment,
  title={Segment anything},
  author={Kirillov, Alexander and Mintun, Eric and Ravi, Nikhila and Mao, Hanzi and Rolland, Chloe and Gustafson, Laura and Xiao, Tete and Whitehead, Spencer and Berg, Alexander C and Lo, Wan-Yen and others},
  booktitle={Proceedings of the IEEE/CVF international conference on computer vision},
  pages={4015--4026},
  year={2023}
}

@inproceedings{lin2014microsoft,
  title={Microsoft coco: Common objects in context},
  author={Lin, Tsung-Yi and Maire, Michael and Belongie, Serge and Hays, James and Perona, Pietro and Ramanan, Deva and Doll{\'a}r, Piotr and Zitnick, C Lawrence},
  booktitle={European conference on computer vision},
  pages={740--755},
  year={2014},
  organization={Springer}
}

@article{meng2025vlm2vec2,
  title={VLM2Vec-V2: Advancing Multimodal Embedding for Videos, Images, and Visual Documents},
  author={Meng, Rui and Jiang, Ziyan and Liu, Ye and Su, Mingyi and Yang, Xinyi and Fu, Yuepeng and Qin, Can and Chen, Zeyuan and Xu, Ran and Xiong, Caiming and others},
  journal={arXiv preprint arXiv:2507.04590},
  year={2025}
}

@article{tschannen2025siglip,
  title={Siglip 2: Multilingual vision-language encoders with improved semantic understanding, localization, and dense features},
  author={Tschannen, Michael and Gritsenko, Alexey and Wang, Xiao and Naeem, Muhammad Ferjad and Alabdulmohsin, Ibrahim and Parthasarathy, Nikhil and Evans, Talfan and Beyer, Lucas and Xia, Ye and Mustafa, Basil and others},
  journal={arXiv preprint arXiv:2502.14786},
  year={2025}
}

@inproceedings{zheng2024dreamlip,
  title={Dreamlip: Language-image pre-training with long captions},
  author={Zheng, Kecheng and Zhang, Yifei and Wu, Wei and Lu, Fan and Ma, Shuailei and Jin, Xin and Chen, Wei and Shen, Yujun},
  booktitle={European Conference on Computer Vision},
  pages={73--90},
  year={2024},
  organization={Springer}
}

@article{wang2022e5,
  title={Text embeddings by weakly-supervised contrastive pre-training},
  author={Wang, Liang and Yang, Nan and Huang, Xiaolong and Jiao, Binxing and Yang, Linjun and Jiang, Daxin and Majumder, Rangan and Wei, Furu},
  journal={arXiv preprint arXiv:2212.03533},
  year={2022}
}

@misc{lee2024nvembed,
      title={NV-Embed: Improved Techniques for Training LLMs as Generalist Embedding Models}, 
      author={Chankyu Lee and Rajarshi Roy and Mengyao Xu and Jonathan Raiman and Mohammad Shoeybi and Bryan Catanzaro and Wei Ping},
      year={2024},
      eprint={2405.17428},
      archivePrefix={arXiv},
      primaryClass={cs.CL}
}

@inproceedings{xie2025fgclip,
      title={FG-CLIP: Fine-Grained Visual and Textual Alignment}, 
      author={Chunyu Xie and Bin Wang and Fanjing Kong and Jincheng Li and Dawei Liang and Gengshen Zhang and Dawei Leng and Yuhui Yin},
      year={2025},
      booktitle=icml
}

@article{jing2024fineclip,
  title={Fineclip: Self-distilled region-based clip for better fine-grained understanding},
  author={Jing, Dong and He, Xiaolong and Luo, Yutian and Fei, Nanyi and Wei, Wei and Zhao, Huiwen and Lu, Zhiwu and others},
  journal={Advances in Neural Information Processing Systems},
  volume={37},
  pages={27896--27918},
  year={2024}
}

@article{kirillov2023segany,
  title={Segment Anything},
  author={Kirillov, Alexander and Mintun, Eric and Ravi, Nikhila and Mao, Hanzi and Rolland, Chloe and Gustafson, Laura and Xiao, Tete and Whitehead, Spencer and Berg, Alexander C. and Lo, Wan-Yen and Doll{\'a}r, Piotr and Girshick, Ross},
  journal={arXiv:2304.02643},
  year={2023}
}

@inproceedings{dong2022m5product,
  title={M5Product: Self-harmonized Contrastive Learning for E-commercial Multi-modal Pretraining},
  author={Dong, Xiao and Zhan, Xunlin and Wu, Yangxin and Wei, Yunchao and Kampffmeyer, Michael C and Wei, Xiaoyong and Lu, Minlong and Wang, Yaowei and Liang, Xiaodan},
  booktitle={2022 IEEE/CVF Conference on Computer Vision and Pattern Recognition (CVPR)},
  pages={21220--21230},
  year={2022},
  organization={IEEE Computer Society}
}

@article{yang2023setofmark,
      title={Set-of-Mark Prompting Unleashes Extraordinary Visual Grounding in GPT-4V}, 
      author={Jianwei Yang and Hao Zhang and Feng Li and Xueyan Zou and Chunyuan Li and Jianfeng Gao},
      journal={arXiv preprint arXiv:2310.11441},
      year={2023},
}

@inproceedings{yang2018hotpotqa,
  title={{HotpotQA}: A Dataset for Diverse, Explainable Multi-hop Question Answering},
  author={Yang, Zhilin and Qi, Peng and Zhang, Saizheng and Bengio, Yoshua and Cohen, William W. and Salakhutdinov, Ruslan and Manning, Christopher D.},
  booktitle={Conference on Empirical Methods in Natural Language Processing ({EMNLP})},
  year={2018}
}

@misc{lin2025pam,
    title={Perceive Anything: Recognize, Explain, Caption, and Segment Anything in Images and Videos}, 
    author={Weifeng Lin and Xinyu Wei and Ruichuan An and Tianhe Ren and Tingwei Chen and Renrui Zhang and Ziyu Guo and Wentao Zhang and Lei Zhang and Hongsheng Li},
    year={2025},
    eprint={2506.05302},
    archivePrefix={arXiv},
    primaryClass={cs.CV}
}

@article{he2018layer-wise,
  title={Layer-Wise Coordination between Encoder and Decoder for Neural Machine Translation},
  author={He, Tianyu and Tan, Xu and Xia, Yingce and He, Di and Qin, Tao and Chen, Zhibo and Liu, Tie-Yan},
  journal={Advances in Neural Information Processing Systems},
  volume={31},
  year={2018}
}

@article{MARCO,
  author    = {Tri Nguyen and
               Mir Rosenberg and
               Xia Song and
               Jianfeng Gao and
               Saurabh Tiwary and
               Rangan Majumder and
               Li Deng},
  title     = {{MS} {MARCO:} {A} Human Generated MAchine Reading COmprehension Dataset},
  journal   = {CoRR},
  volume    = {abs/1611.09268},
  year      = {2016},
  url       = {http://arxiv.org/abs/1611.09268},
  archivePrefix = {arXiv},
  eprint    = {1611.09268},
  bibsource = {dblp computer science bibliography, https://dblp.org}
}

@misc{yu2025vpt,
      title={Introducing Visual Perception Token into Multimodal Large Language Model}, 
      author={Runpeng Yu and Xinyin Ma and Xinchao Wang},
      year={2025},
      eprint={2502.17425},
      archivePrefix={arXiv},
}

@misc{jiao2024imgdiffcontrastivedatasynthesis,
      title={Img-Diff: Contrastive Data Synthesis for Multimodal Large Language Models}, 
      author={Qirui Jiao and Daoyuan Chen and Yilun Huang and Bolin Ding and Yaliang Li and Ying Shen},
      year={2024},
      eprint={2408.04594},
      archivePrefix={arXiv},
      primaryClass={cs.CV},
      url={https://arxiv.org/abs/2408.04594}, 
}

@inproceedings{vismin2024,
    title={VisMin: Visual Minimal-Change Understanding},
    author={Awal, Rabiul and Ahmadi, Saba and Zhang, Le and Agrawal, Aishwarya},
    year={2024},
    booktitle=nips
}

@misc{kakaobrain2022coyo-700m,
  title         = {COYO-700M: Image-Text Pair Dataset},
  author        = {Byeon, Minwoo and Park, Beomhee and Kim, Haecheon and Lee, Sungjun and Baek, Woonhyuk and Kim, Saehoon},
  year          = {2022},
  howpublished  = {\url{https://github.com/kakaobrain/coyo-dataset}},
}

@article{zhang2024gme,
  title={GME: Improving Universal Multimodal Retrieval by Multimodal LLMs},
  author={Zhang, Xin and Zhang, Yanzhao and Xie, Wen and Li, Mingxin and Dai, Ziqi and Long, Dingkun and Xie, Pengjun and Zhang, Meishan and Li, Wenjie and Zhang, Min},
  journal={arXiv preprint arXiv:2412.16855},
  year={2024}
}

@article{lian2025dam,
  title={Describe anything: Detailed localized image and video captioning},
  author={Lian, Long and Ding, Yifan and Ge, Yunhao and Liu, Sifei and Mao, Hanzi and Li, Boyi and Pavone, Marco and Liu, Ming-Yu and Darrell, Trevor and Yala, Adam and others},
  journal={arXiv preprint arXiv:2504.16072},
  year={2025}
}

@inproceedings{zhang2024long,
  title={Long-clip: Unlocking the long-text capability of clip},
  author={Zhang, Beichen and Zhang, Pan and Dong, Xiaoyi and Zang, Yuhang and Wang, Jiaqi},
  booktitle={European Conference on Computer Vision},
  year={2024},
}

@article{sun2024eva,
  title={Eva-clip-18b: Scaling clip to 18 billion parameters},
  author={Sun, Quan and Wang, Jinsheng and Yu, Qiying and Cui, Yufeng and Zhang, Fan and Zhang, Xiaosong and Wang, Xinlong},
  journal={arXiv preprint arXiv:2402.04252},
  year={2024}
}

@inproceedings{jin2023eclip,
  title={Learning instance-level representation for large-scale multi-modal pretraining in e-commerce},
  author={Jin, Yang and Li, Yongzhi and Yuan, Zehuan and Mu, Yadong},
  booktitle={Proceedings of the IEEE/CVF Conference on Computer Vision and Pattern Recognition},
  pages={11060--11069},
  year={2023}
}

@article{oord2018representation,
  title={Representation learning with contrastive predictive coding},
  author={Oord, Aaron van den and Li, Yazhe and Vinyals, Oriol},
  journal={arXiv preprint arXiv:1807.03748},
  year={2018}
}

@inproceedings{gao2021simcse,
  title={SimCSE: Simple Contrastive Learning of Sentence Embeddings},
  author={Gao, Tianyu and Yao, Xingcheng and Chen, Danqi},
  booktitle=EMNLP,
  year={2021}
}

@article{wang2024qwen2,
  title={Qwen2-VL: Enhancing Vision-Language Model's Perception of the World at Any Resolution},
  author={Wang, Peng and Bai, Shuai and Tan, Sinan and Wang, Shijie and Fan, Zhihao and Bai, Jinze and Chen, Keqin and Liu, Xuejing and Wang, Jialin and Ge, Wenbin and others},
  journal={arXiv preprint arXiv:2409.12191},
  year={2024}
}

@inproceedings{jia2021scaling,
  title={Scaling up visual and vision-language representation learning with noisy text supervision},
  author={Jia, Chao and Yang, Yinfei and Xia, Ye and Chen, Yi-Ting and Parekh, Zarana and Pham, Hieu and Le, Quoc and Sung, Yun-Hsuan and Li, Zhen and Duerig, Tom},
  booktitle=icml,
  year={2021}
}

@inproceedings{zhai2023sigmoid,
  title={Sigmoid loss for language image pre-training},
  author={Zhai, Xiaohua and Mustafa, Basil and Kolesnikov, Alexander and Beyer, Lucas},
  booktitle=iccv,
  year={2023}
}

@inproceedings{li2023blip,
  title={Blip-2: Bootstrapping language-image pre-training with frozen image encoders and large language models},
  author={Li, Junnan and Li, Dongxu and Savarese, Silvio and Hoi, Steven},
  booktitle=icml,
  year={2023},
}

@article{jiang2024vlm2vec,
  title={VLM2Vec: Training Vision-Language Models for Massive Multimodal Embedding Tasks},
  author={Jiang, Ziyan and Meng, Rui and Yang, Xinyi and Yavuz, Semih and Zhou, Yingbo and Chen, Wenhu},
  journal={arXiv preprint arXiv:2410.05160},
  year={2024}
}

@inproceedings{lin2023fine,
  title={Fine-grained late-interaction multi-modal retrieval for retrieval augmented visual question answering},
  author={Lin, Weizhe and Chen, Jinghong and Mei, Jingbiao and Coca, Alexandru and Byrne, Bill},
  booktitle=nips,
  year={2023}
}

@article{jian2025rzenembed,
  title={Rzenembed: Towards comprehensive multimodal retrieval},
  author={Jian, Weijian and Zhang, Yajun and Liang, Dawei and Xie, Chunyu and He, Yixiao and Leng, Dawei and Yin, Yuhui},
  journal={arXiv preprint arXiv:2510.27350},
  year={2025}
}

@article{lee2023medical,
  title={Region-based contrastive pretraining for medical image retrieval with anatomic query},
  author={Lee, Ho Hin and Santamaria-Pang, Alberto and Merkow, Jameson and Oktay, Ozan and P{\'e}rez-Garc{\'\i}a, Fernando and Alvarez-Valle, Javier and Tarapov, Ivan},
  journal={arXiv preprint arXiv:2305.05598},
  year={2023}
}

@misc{zhang2025gpt4roi,
      title={GPT4RoI: Instruction Tuning Large Language Model on Region-of-Interest}, 
      author={Shilong Zhang and Peize Sun and Shoufa Chen and Min Xiao and Wenqi Shao and Wenwei Zhang and Yu Liu and Kai Chen and Ping Luo},
      year={2025},
      eprint={2307.03601},
      archivePrefix={arXiv},
      primaryClass={cs.CV},
      url={https://arxiv.org/abs/2307.03601}, 
}

@inproceedings{guo2024regiongpt,
  title={Regiongpt: Towards region understanding vision language model},
  author={Guo, Qiushan and De Mello, Shalini and Yin, Hongxu and Byeon, Wonmin and Cheung, Ka Chun and Yu, Yizhou and Luo, Ping and Liu, Sifei},
  booktitle={Proceedings of the IEEE/CVF Conference on Computer Vision and Pattern Recognition},
  pages={13796--13806},
  year={2024}
}

@article{wang2025grasp,
  title={Grasp Any Region: Towards Precise, Contextual Pixel Understanding for Multimodal LLMs},
  author={Wang, Haochen and Wang, Yuhao and Zhang, Tao and Zhou, Yikang and Li, Yanwei and Wang, Jiacong and Zheng, Jiani and Tian, Ye and Meng, Jiahao and Huang, Zilong and others},
  journal={arXiv preprint arXiv:2510.18876},
  year={2025}
}

@misc{faysse2025colpaliefficientdocumentretrieval,
      title={ColPali: Efficient Document Retrieval with Vision Language Models}, 
      author={Manuel Faysse and Hugues Sibille and Tony Wu and Bilel Omrani and Gautier Viaud and Céline Hudelot and Pierre Colombo},
      year={2025},
      eprint={2407.01449},
      archivePrefix={arXiv},
      primaryClass={cs.IR},
      url={https://arxiv.org/abs/2407.01449}, 
}

@article{chen2025mmE5,
  title={mmE5: Improving Multimodal Multilingual Embeddings via High-quality Synthetic Data},
  author={Chen, Haonan and Wang, Liang and Yang, Nan and Zhu, Yutao and Zhao, Ziliang and Wei, Furu and Dou, Zhicheng},
  journal={arXiv preprint arXiv:2502.08468},
  year={2025}
}

@inproceedings{li2021unimo,
  title={Unimo: Towards unified-modal understanding and generation via cross-modal contrastive learning},
  author={Li, Wei and Gao, Can and Niu, Guocheng and Xiao, Xinyan and Liu, Hao and Liu, Jiachen and Wu, Hua and Wang, Haifeng},
  booktitle={Proceedings of the 59th Annual Meeting of the Association for Computational Linguistics and the 11th International Joint Conference on Natural Language Processing (Volume 1: Long Papers)},
  pages={2592--2607},
  year={2021}
}

@article{yao2021filip,
  title={Filip: Fine-grained interactive language-image pre-training},
  author={Yao, Lewei and Huang, Runhui and Hou, Lu and Lu, Guansong and Niu, Minzhe and Xu, Hang and Liang, Xiaodan and Li, Zhenguo and Jiang, Xin and Xu, Chunjing},
  journal={arXiv preprint arXiv:2111.07783},
  year={2021}
}

@inproceedings{zhang2018visual,
  title={Visual search at alibaba},
  author={Zhang, Yanhao and Pan, Pan and Zheng, Yun and Zhao, Kang and Zhang, Yingya and Ren, Xiaofeng and Jin, Rong},
  booktitle={Proceedings of the 24th ACM SIGKDD international conference on knowledge discovery \& data mining},
  pages={993--1001},
  year={2018}
}

@misc{kim2026pixelgroundedretrievalknowledgeablelarge,
      title={Pixel-Grounded Retrieval for Knowledgeable Large Multimodal Models}, 
      author={Jeonghwan Kim and Renjie Tao and Sanat Sharma and Jiaqi Wang and Kai Sun and Zhaojiang Lin and Seungwhan Moon and Lambert Mathias and Anuj Kumar and Heng Ji and Xin Luna Dong},
      year={2026},
      eprint={2601.19060},
      archivePrefix={arXiv},
      primaryClass={cs.CV},
      url={https://arxiv.org/abs/2601.19060}, 
}

@article{DeepFashion2,
  author = {Yuying Ge and Ruimao Zhang and Lingyun Wu and Xiaogang Wang and Xiaoou Tang and Ping Luo},
  title={A Versatile Benchmark for Detection, Pose Estimation, Segmentation and Re-Identification of Clothing Images},
  journal={CVPR},
  year={2019}
}

@misc{ilias,
      title={ILIAS: Instance-Level Image retrieval At Scale}, 
      author={Giorgos Kordopatis-Zilos and Vladan Stojnić and Anna Manko and Pavel Šuma and Nikolaos-Antonios Ypsilantis and Nikos Efthymiadis and Zakaria Laskar and Jiří Matas and Ondřej Chum and Giorgos Tolias},
      year={2025},
      eprint={2502.11748},
      archivePrefix={arXiv},
      primaryClass={cs.CV},
      url={https://arxiv.org/abs/2502.11748}, 
}

@inproceedings{radenovic2018revisitingoxford,
  title={Revisiting oxford and paris: Large-scale image retrieval benchmarking},
  author={Radenovi{\'c}, Filip and Iscen, Ahmet and Tolias, Giorgos and Avrithis, Yannis and Chum, Ond{\v{r}}ej},
  booktitle={Proceedings of the IEEE conference on computer vision and pattern recognition},
  pages={5706--5715},
  year={2018}
}
